# Development and Evaluation of Ensemble Learning-based Environmental Methane Detection and Intensity Prediction Models

Reek Majumder*
Glenn Department of Civil Engineering
Clemson University, Clemson, SC 29631, USA
Email: rmajumd@clemson.edu

Jacquan Pollard
Glenn Department of Civil Engineering
Clemson University, Clemson, SC 29631, USA
Email: jacquap@clemson.edu

M Sabbir Salek*
Glenn Department of Civil Engineering
Clemson University, Clemson, SC 29631, USA
Email: msalek@clemson.edu

David Werth, PhD
Savannah River National Laboratory, Aiken, SC 29831
Email: david.werth@srnl.doe.gov

Gurcan Comert, PhD
Comp. Sci., Phy., and Engineering Department, Benedict College
1600 Harden Street, Columbia, SC 29204.
Email: gurcan.comert@Benedict.edu

Adrian Gale, PhD
Comp. Sci., Phy., and Engineering Department, Benedict College
1600 Harden Street, Columbia, SC 29204
Email: adrian.gale@benedict.edu

Sakib Mahmud Khan, PhD
Glenn Department of Civil Engineering
Clemson University, Clemson, SC 29631, USA
Email: sakibk@clemson.edu

Samuel Darko, PhD
School of Arts and Sciences
Florida Memorial University, Miami Gardens, FL 33054, USA
Email: samuel.darko@fmuniv.edu

Mashrur Chowdhury, PhD
Department of Civil Engineering
Clemson University, Clemson, SC 29631, USA
Email: mac@clemson.edu

*Corresponding Author

*Abstract*--The environmental impacts of global warming driven by methane ($CH_4$) emissions have catalyzed significant research initiatives in developing novel technologies that enable proactive and rapid detection of $CH_4$. Several data-driven machine learning (ML) models were tested to determine how well they identified fugitive $CH_4$ and its related intensity in the affected areas. Various meteorological characteristics, including wind speed, temperature, pressure, relative humidity, water vapor, and heat flux, were included in the simulation. We used the ensemble learning method to determine the best-performing weighted ensemble ML models built upon several weaker lower-layer ML models to (i) detect the presence of $CH_4$ as a classification problem and (ii) predict the intensity of $CH_4$ as a regression problem. The classification model performance for $CH_4$ detection was evaluated using accuracy, F1 score, Matthew's Correlation Coefficient (MCC), and the area under the receiver operating characteristic curve (AUC ROC), with the output of the top-performing model being 97.2%, 0.972, 0.945 and 0.995, respectively. The R2 score was used to evaluate the regression model performance for $CH_4$ intensity prediction, with the R2 score of the best-performing model being 0.858. The ML models developed in this study for fugitive $CH_4$ detection and intensity prediction can be used with fixed environmental sensors deployed on the ground or with sensors mounted on unmanned aerial vehicles (UAVs) for mobile detection.

I. INTRODUCTION

Methane ($CH_4$) emissions have garnered considerable research interest in recent years. Governmental organizations like the Environmental Protection Agency (EPA) have attempted to improve inaccurate estimates of industrial $CH_4$ leakage [1]. Methane is a powerful greenhouse gas (GHG) that contributes to global warming. Its impact on the environment has been found to possess far greater consequences on global climate systems over shorter periods than other GHGs such as carbon dioxide ($CO_2$), nitrous oxide ($N_2O$), chlorofluorocarbons (CFC), and others [2]. The significance of this finding is reflected in the global warming potential (GWP), a metric, that measures the global warming impact of various GHGs relative to 1 ton of $CO_2$. With a GWP of methane 27-30 over a 100-year timescale,[3], it is shown that although $CH_4$ is a short-lived GHG, it is far more efficient at absorbing radiation than $CO_2$ and contributes significantly to global warming [4]. This finding indicates that methane's propensity for absorbing and trapping radiation in the atmosphere negatively impacts the environment substantially more than $CO_2$ over a 20-year timeframe.

The largest source of $CH_4$ emissions in the U.S. is attributed to the energy sector by processes related to the production, extraction, and transportation of fossil fuels [5]. As the primary constituent in natural gas, $CH_4$ may enter the environment through cracks and fissures in degraded or defective pipeline infrastructure which spans over 2.6 million miles. These pipeline networks deliver 25 percent of the energy consumed by the U.S. population for various uses, such as fueling automobiles and regulating temperatures inside residential and commercial buildings [6]. However, the traversing of oil and natural gas (ONG) infrastructure through major urban areas compounds the dangers of fugitive $CH_4$ emissions. Consequently, it raises negative implications for human health and environmental impacts due to $CH_4$ emissions being devoid of color and odor and possessing a highly volatile nature [7]. The increased focus on $CH_4$ emissions caused by ONG distribution systems has been motivated by the substantial benefits of employing early detection technologies for $CH_4$ emissions. Manual measurement methods for the early detection of methane suffer from various limitations. These include limited sampling coverage that may not account for localized emissions, delayed detection, a lack of real-time monitoring, and risks to personnel safety. Moreover, the considerable possibility of human error and a lack of data analysis and integration from multiple sensors make it difficult to implement early-warning programs. According to recent studies [8] [9], the average time to perform mitigating procedures for fugitive $CH_4$ emissions will be significantly reduced by employing large-scale autonomous environmental emission detection systems requiring minimal human involvement.

Statistical methods and data analysis techniques employed by artificial intelligence and machine learning (ML) algorithms provide the ability to extract useful information from complex data-driven analytical tasks. Recent studies conducted in environmental monitoring applications have employed novel approaches using the intelligent decision-making capabilities of ML algorithms to guide the efficient use of limited resources [10]. In an environmental emissions application, ML algorithms can detect the presence of harmful pollutants or contaminants by processing atmospheric meteorological data. A heterogeneous environmental sensor network consisting of fixed and mobile sensors backed by ML can improve environmental surveillance of fugitive $CH_4$ emissions by being distributed across multiple sensor edges and geographical areas for large-scale detection, mitigation, and reduced operator response times. This paper aims to develop and assess the performance of the Ensemble Learning model developed with several data-driven ML models to achieve an autonomous and efficient detection of fugitive $CH_4$ emissions with minimal false alarms.

The rest of this paper is organized as follows: the review of the literature is covered in Section 2. In Section 3, we describe the dataset used in this study. Section 4 details the framework of our proposed fugitive $CH_4$ environmental emissions detection system. The experimental analysis and a discussion of the results obtained in our study are presented in Section 5. Section 6 discusses how the emission detection system can be scaled into a distributed $CH_4$ environmental emissions detection system having multiple sensor edges and geographical areas for large-scale detection, mitigation, and reduced operator response times. Section 7 concludes this paper and presents some remarks for future research initiatives.

II. LITERATURE REVIEW

Many studies in the literature have investigated various strategies to implement large-scale environmental monitoring and mitigation of fugitive $CH_4$ emissions. This section provides a brief discussion of previous research focused on large-scale environmental assessments, their limitations, and a brief discussion of the gaps in the knowledge that persist.

In a comparative review conducted in 2021, Sun et al., investigated how $CH_4$ emissions estimation technologies, data analysis, and probabilistic tools for emission uncertainty estimation unite to form an efficient system of $CH_4$ management for the ONG energy sector [11]. The authors found that spatial scale was a primary factor in selecting $CH_4$ emission measurement technology where bottom-up or top-down methods were applied to calculate emission rates from $CH_4$ concentrations. Bu et al. addressed the leakage and diffusion characteristics of fugitive $CH_4$ emissions in utility tunnels using a simulation-based approach under various working conditions [12]. They achieved an average error of 8.29% using a methane invasion distance (MID) prediction equation to accurately predict the location of fugitive $CH_4$ emissions under various leakage times for optimal placement of utility tunnel alarm devices. Fleming et al. conducted a field-based case study that examined $CO_2$ and $CH_4$ efflux measurements over two weeks around a petroleum well in Alberta, Canada, using automated dynamic closed chambers[13]. The authors concluded that quantifying the efflux rate of $CO_2$ and $CH_4$ was challenging due to spatiotemporal emission variability and suggested that accounts for spatiotemporal variability must be considered for accurate fugitive $CH_4$ migration estimations. Bezyk et al. used the static chamber method to investigate the factors inducing $CH_4$ uptake for three different land uses in Wroclaw, Poland [14]. The findings from their study showed that a threshold for soil temperature and moisture in urban environments acts as a sink for fugitive $CH_4$ emissions. Okorn et al. used innovative sensors to measure gaseous pollutants, such as fugitive $CH_4$ emissions, $CO_2$, carbon monoxide, etc., in three different urban areas in Los Angeles [15]. The authors used a multi-sensor linear regression calibration model and achieved an improvement of 16% coefficient of determination and a 22% reduction in the root mean square error.

Hahnel et al. applied a deep learning framework for air pollution monitoring and forecasting to extend model training across different domains [16], whereas Eldakhly et al. investigated several methods to determine the best-performing ML model to forecast air pollution in the Greater Cairo Metropolitan Area (GCMA) in Egypt [17]. The authors reported that higher model accuracy was achieved by modifying the support vector regression (SVR) algorithm with the chance weight based on chance theory but suffered long execution run times. Wang et al. developed an unsupervised ML framework for sensor optimization around active operation sites that combines multiple data sources, including oil and gas infrastructure data, historical methane leak rate distribution, topographical data, and meteorological data [4]. The proposed framework achieved an 87.9% success rate in detecting $CH_4$ leak sources, whereas the baseline model detected 82.8% of sources responsible for fugitive $CH_4$ emissions. In Travis et al., an artificial neural network (ANN) was developed using input information as measured, by sonic anemometer wind velocities and $CH_4$ sensor time series. This allowed the authors to infer the location and release rate of a gas leak [16]. The authors found that the ANN failed to detect very low leak rates and that parameters, such as wind speed and direction, can negatively impact the model's performance. Wang et al. considered the problem of methane leak size classification using a video classification task. This enabled the authors to develop deep learning models to classify videos by leak volume [7]. The authors compared three deep learning algorithms and achieved leak and non-leak detection success rates of nearly 100%.

III. MATERIALS AND ANALYSIS PRE-REQUISITE

*A. Dataset*

The "data" used in our study was generated using a comprehensive mesoscale meteorological modeling system, the Regional Atmospheric Modeling System [17], also referred to as RAMS, which is used by the

Savannah River National Laboratory (SRNL) for weather forecasting and airborne dispersion to simulate toxic release. The model solves the explicit dynamic equations for different meteorological parameters on a three-dimensional grid for a given location and time to simulate atmospheric dynamics, thermodynamics, precipitation, and land surface processes. The RAMS model configuration used in this study is similar to that described in Werth and Buckley [18], but using a finer grid. It was set to run over a 125-meter grid on a 7.6 x 7.6-kilometer domain. The model ran for six hours, starting on February 10th, 2022, at 1200 UTC and ending on February 10th, 2022, at 1800 UTC. During the runtime, data was saved every six minutes. The model data covers a latitude from 33.216 °N to 33.284 °N and a longitude from -81.69 °W to -81.61 °W.

The Hybrid Single-Particle Lagrangian Integrated Trajectory model (HYSPLIT, Stein et al. [19]) is widely used for trajectory mapping and dispersion simulation models of atmospheric releases, which require meteorological data as input. Meteorological data collected from the RAMS model were used in a subsequent run of the HYSPLIT model to simulate an airborne tracer's downwind motion and mixing and to predict the tracer concentrations at various locations. The weather and concentration data were organized for each grid location at six-minute intervals.

The simulation was centered over the Savannah River Site (SRS). The SRS comprises approximately 310 square miles and lies alongside the Savannah River. The landscape is primarily forested with pine trees ranging from 10 to 20 meters in height, and industrial facilities are strategically clustered in separate areas connected by roads, which includes multiple buildings of varying sizes. The simulated release point was at 33.25°N and 81.65°W. The topography consists of rolling hills, gradually descending from an elevation of 128 meters above sea level in the northeast to 30 meters above sea level in the southwest. The lowest points are found along the floodplain bordering the river at the southwest boundary, where several on-site creeks drain. During the 6-hour simulation, the wind direction transitions from midway between west and northwest (WNW) to southwest (SW). Assuming a large but plausible release (147 metric tons per hour) our tracer concentration values from the HYSPLIT model are on the order of 0.2 to 2.0 parts per million by volume (ppm-V).

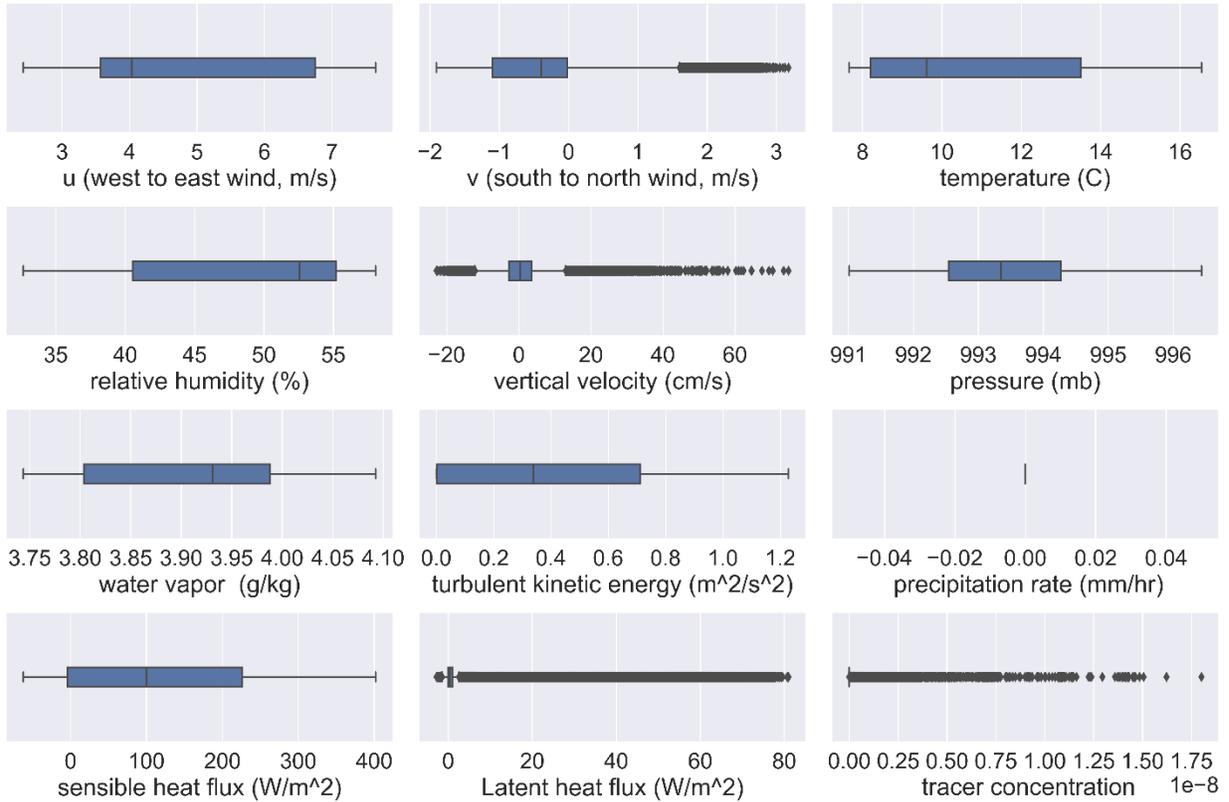

Fig. 1. Boxplots of the meteorological parameters obtained from the RAMS model and tracer concentration from the HYSPLIT model

The simulation dataset consists of 301,340 entries with one timing variable (i.e., time), two location variables (i.e., latitude and longitude), and twelve meteorological parameters, such as temperature, relative humidity, pressure, water vapor, turbulent kinetic energy, precipitation rate, sensible heat flux, latent heat flux, and west to east wind velocity, south to north wind velocity, and vertical wind velocity and tracer concentration. Figure 1 presents descriptive statistics of the various meteorological parameters obtained from the RAMS model and Tracer Concentration obtained from the HYSPLIT model.

### B. Data Preprocessing

Among the 301,340 entries in our dataset, only 8,850 entries are associated with $CH_4$ leakage (i.e., Tracer concentration > 0), whereas it has 292,490 entries with no $CH_4$ (i.e., Tracer concentration = 0). Therefore, we used random under-sampling to extract 8,850 entries with no $CH_4$ leakage to get a balanced dataset. Our balanced dataset consists of 17,700 entries, including an equal number of entries with $CH_4$ leakage and no $CH_4$ leakage. Next, we removed the timing and location variables from the balanced dataset, as the ML models should only depend on the meteorological parameters. Moreover, from Figure 1, we see precipitation rate has a constant value of zero for all the entries of the entire dataset, so we dropped this parameter. Then, we have a cleaned dataset of 17,700 entries with eleven meteorological parameters. Next, we introduced a new binary label Leakage, where leakage equals class 1 if the tracer concentration is greater than zero otherwise to class 0. We removed Tracer concentration from the dataset, used Leakage as the target binary label, and the ten meteorological parameters as the input features for the classification binary label, and the ten meteorological parameters as the input features.

For the regression model we only considered entries with Tracer concentration > 0, where Tracer concentration served as the target variable, and the ten meteorological parameters served as the input features. For both classification and regression analyses, we used 4:1 random splitting to generate train and test datasets. Since the different meteorological parameters in our dataset have different ranges of values (as seen in Figure 1) and our target variable for regression, i.e., Tracer concentration, holds extremely small values compared to the other meteorological parameters. We scaled the parameters in our training and testing datasets for regression analysis based on the equation, scale value = $(x - \mu)/\sigma$ where $\mu$ and $\sigma$ denote the mean and the standard deviation of a given meteorological parameters, respectively. Figure 2 presents all the steps of data preprocessing for a better understanding.

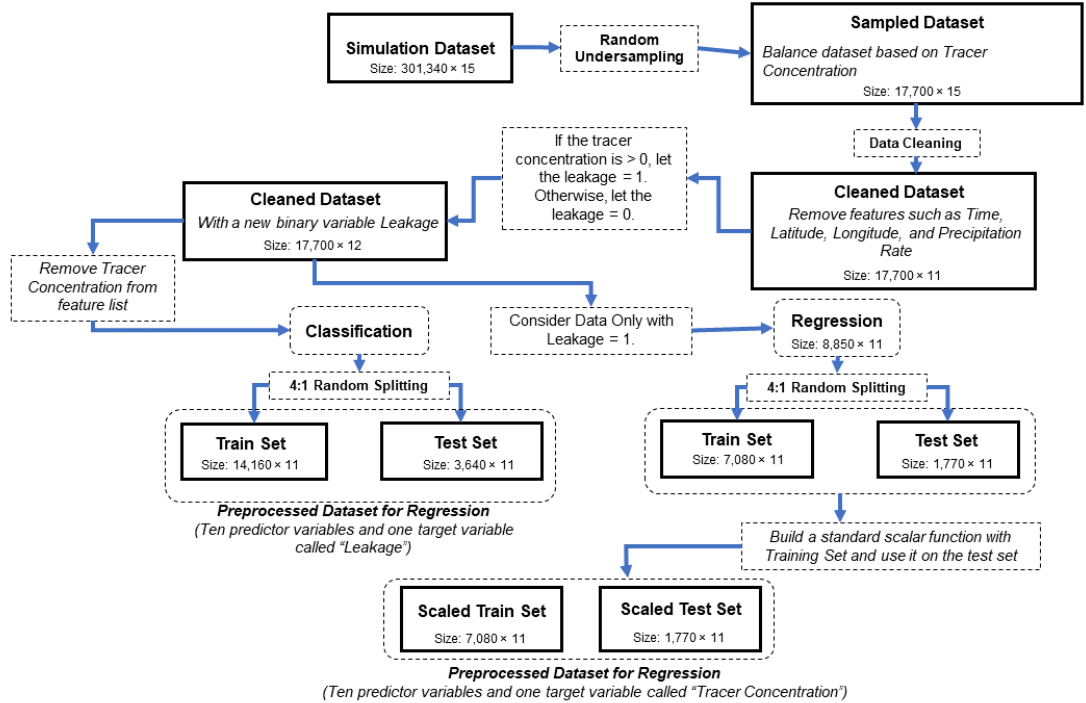

Fig. 2. Data preprocessing and analysis step

## C. Ensemble Learning Models and Hyperparameters Tuning

Adjustable parameters used during ML model development are called hyperparameters, which regulate how the model trains, prevent overfitting on training data and enhance the model's performance on unknown data. The hyperparameters are tuned for regression and classification tasks to maximize an objective function. For our analysis we have chosen accuracy as an objective function for the classification model and $R^2$ score for the regression models. We have automated our hyperparameter tuning task using Bayesian optimization and hyperband (BOHB) [20] from the ray tune [21] library to get the optimal set of hyperparameters for our ML models.

*Bayesian Optimization and Hyperband (BOHB):* It is the state-of-the-art algorithm for hyperparameter optimization that combines the efficiency of hyperband's [22] successive halving strategy for resource allocation with global optimization capability of Bayesian optimization[23] by maintaining a probabilistic model of the objective function based on observed performance. It efficiently navigates the hyperparameter search space iteratively by refining its understanding of promising configurations and allocating resources judiciously.

For our study, we have mainly used three types of models (1) tree-based models, (2) gradient boosting models, and (3) K-nearest neighbors.

*Tree-Based Models:* Among the tree-based models we employed in our research were Extra Trees (ET) and Random Forest (RF), constructed using several decision trees. The maximum depth of the tree (max_depth), the minimum number of samples needed to split an internal node (min_samples_split), the minimum number of samples needed to be at a leaf node (min_samples_leaf), and the total number of trees utilized by these models (n_estimators) are the critical parameters for these models. The criterion (criterion) parameter, which contains options like "gini" for Gini impurity and "entropy" for information gain, specifies the quality metric for node splitting.

*Gradient boosting models:* In our investigation, we employed various gradient boosting models, including the Light Gradient Boosting Model (LGBM), the Extreme Gradient Boosting Model (XGBoost), and the categorical boosting (CatBoost) model. The number of boosting rounds (num_boost_round) and learning rate (learning_rate) are critical hyperparameters for these models.

*K-Nearest Neighbor:* The primary hyperparameters in K-Nearest Neighbor (KNN) used are "weights" which affects how much surrounding data points contribute to predictions.

## IV. DISCUSSION ON METHODS AND RESULTS

RAMS tools are actively used in SRNL to gather meteorological parameters and as input to various airborne dispersion models for detecting toxic gases. The HYSPLIT is a physics-based model widely used in gaseous dispersion modeling. These models encounter limitations like scalability, and computational requirements and often fail to map changes in atmospheric conditions. Furthermore, the HYSPLIT primarily relies on meteorological data; it is more difficult to integrate it with other data sources, like satellite imagery or real-time sensor data, which would limit its ability to adapt to changing atmospheric conditions.

The ML-based ensemble learning models present a viable solution for these shortcomings. The ensemble models are well-suited for real-time or near-real-time applications because, once trained, they commonly demonstrate lower computing load and inference time. Additionally, the ensemble learning models allow for the smooth integration of many different data sources, offering a more scalable and flexible method than the gas dispersion models. In addition to their capacity to capture non-linear interactions and manage uncertainty through ensemble aggregation of features, they are useful substitutes for situations involving complicated and dynamic gas dispersion patterns and atmospheric conditions.

In this study, we have used meteorological parameters as an input to ensemble learning models to detect $CH_4$ emissions. However, to obtain ground truth for training and testing the models, we have used the output from the HYSPLIT, a gaseous dispersion model.

Methane Detection Using Ensemble Learning-based Classification Model

We formulated the $CH_4$ detection task as a classification problem by introducing a binary target label Leakage. We used the ensemble learning method to develop and determine the best-performing model for this binary classification problem. The final best-performing model is built upon multiple classifier models using weighted averaging. The underlying classifier models are base layer (Ly1) models, which were directly trained on the input features (i.e., ten meteorological parameters). These models include various configurations of the light gradient boosting method (LGBM), extreme gradient boosting method (XGBoost), random forest (RF), extra trees (ET), k-nearest neighbors (KNN), and category boosting (CatBoost). We used the bootstrap aggregating (bagging) ensemble learning technique to train our base layer models using five-fold cross-validation. Next, weighted averaging was used to determine the optimum combination of the base layer models, resulting in the best-performing model in the output layer (Ly2). Our final best-performing classifier model is a weighted average ensemble of five base layer models (Figure 3). We used an open-source ML library called AutoGluon developed by Amazon [24] to train all our ensemble models.

For performance evaluation of the classifier models, we used four performance metrics, (i) accuracy (ii) F1-Score, (iii) Mathew's Correlation Coefficient (MCC), and (iv) area under the receiver operating characteristics curve (AUC ROC). The performance metrics were determined using the following equations,

$$\text{ACCURACY} = \frac{TP+TN}{TP+FP+TN+FN}$$

$$F1 = \frac{TP}{TP+(FP+FN/2)}$$

$$\text{MCC} = \frac{((TP \times TN)-(FP \times FN))}{\sqrt{(TP+FP)\times(TP+FN)\times(TN+FP)\times(TN+FN)}}$$

$$\text{AUC ROC} = \int_0^1 \frac{TP}{TP+TN} d(\frac{FP}{FP+FN})$$

where, TP, TN, FP, and FN denote true positive, true negative, false positive, and false negative, respectively. Table 1 presents the performance of our base layer models and the final best-performing classifier model. The models presented in Table I show the performance of all the models used in our study. Our final ensemble model achieved an accuracy of 97.2% on the test dataset and the other performance metrics, such as F1 score, Precision, Recall, MCC, and AUC ROC are 0.972, 0.949, 0.997, 0.945, and 0.995. Hyperparameters for our best models (Weighted_Ensemble_L2) and all the contributing base models are presented in Table II.

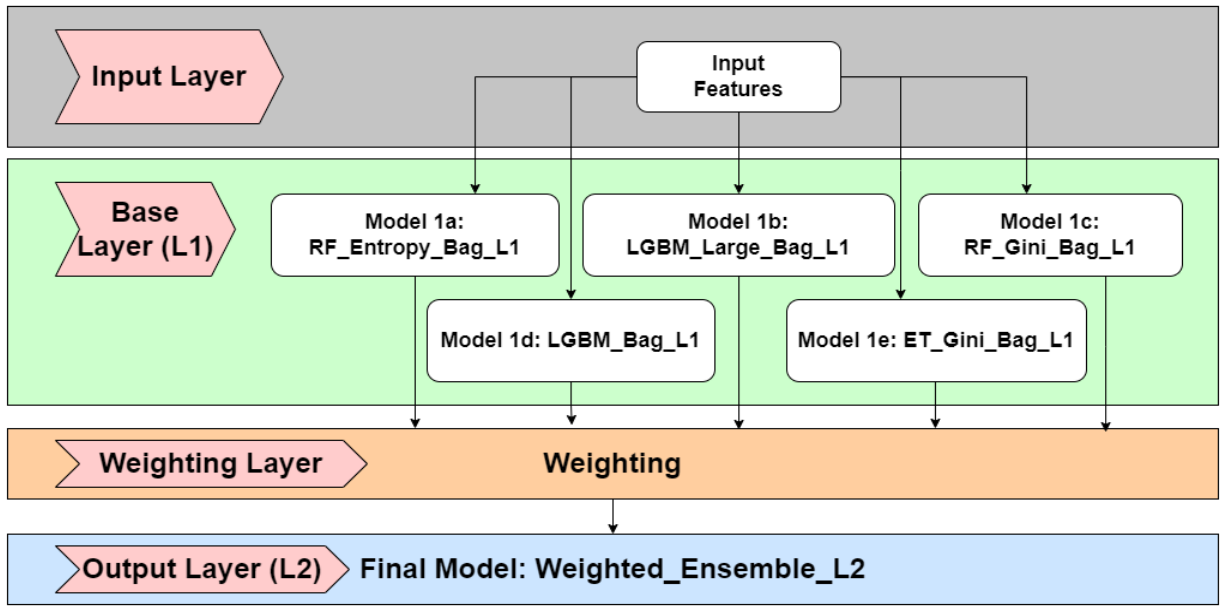

Fig. 3. Steps and the weaker classifier models involved in generating the best-performing classifier model using ensemble learning

TABLE I. PERFORMANCE OF CLASSIFICATION MODELS ON THE TEST SET

| Model | Performance Metric | | | | | |
|---|---|---|---|---|---|---|
| | Accuracy (%) | F1-Score | AUC_ROC | Precision | Recall | MCC |
| WeightedEnsemble_L2 | 97.2 | 0.972 | 0.995 | 0.949 | 0.997 | 0.945 |
| LGBM_Large_BAG_L1 | 97.2 | 0.972 | 0.993 | 0.950 | 0.996 | 0.945 |

| Model | Performance Metric | | | | | |
|---|---|---|---|---|---|---|
| | Accuracy (%) | F1-Score | AUC_ROC | Precision | Recall | MCC |
| LGBM_BAG_L1 | 97.1 | 0.971 | 0.993 | 0.947 | 0.997 | 0.943 |
| XGBoost_BAG_L1 | 96.9 | 0.970 | 0.993 | 0.944 | 0.997 | 0.939 |
| ET_Gini_BAG_L1 | 96.7 | 0.968 | 0.995 | 0.940 | 0.997 | 0.936 |
| RF_Gini_BAG_L1 | 96.7 | 0.968 | 0.995 | 0.943 | 0.994 | 0.935 |
| LGBM_XT_BAG_L1 | 96.7 | 0.967 | 0.991 | 0.941 | 0.995 | 0.935 |
| RF_Entropy_BAG_L1 | 96.6 | 0.967 | 0.995 | 0.942 | 0.994 | 0.934 |
| ET_Entropy_BAG_L1 | 96.6 | 0.967 | 0.995 | 0.938 | 0.997 | 0.934 |
| CatBoost_BAG_L1 | 96.4 | 0.965 | 0.991 | 0.935 | 0.997 | 0.930 |
| KNN_Distance_BAG_L1 | 87.6 | 0.885 | 0.933 | 0.820 | 0.961 | 0.763 |
| KNN_Uniform_BAG_L1 | 87.0 | 0.879 | 0.927 | 0.816 | 0.952 | 0.750 |

TABLE II. HYPERPARAMETER SETTINGS OF THE FINAL MODEL FOR CLASSIFICATION IN THIS STUDY

| Layer | Model | Hyperparameter |
|---|---|---|
| Output Layer (L2) | WeightedEnsemble_L2 | Ensemble Size: 27, RF_Entropy_BAG_L1, RF_Gini_BAG_L1, LGBM_Large_BAG_L1, ET_Gini_BAG_L1, LGBM_BAG_L1 |
| Base Layer (L1) | LGBM_Large_BAG_L1 | Learning Rate: 0.03, Number of Leaves: 128, Feature Fraction: 0.9, Minimum data in leaf: 5, Number of Boost Round: 508 |
| | LGBM_BAG_L1 | Learning Rate: 0.05, Number of Boost Round: 900 |
| | ET_Gini_BAG_L1 | Number of Estimators: 300, Maximum Leaf Nodes: 15000, Criterion: Gini |
| | RF_Gini_BAG_L1 | Number of Estimators: 300, Maximum Leaf Nodes: 15000, Criterion: Gini |
| | RF_Entropy_BAG_L1 | Number of Estimators: 300, Maximum Leaf Nodes: 15000, Criterion: Entropy |

A. *Methane Intensity Prediction Using Ensemble Learning-based Regression Model*

Our $CH_4$ intensity prediction task aims to predict the intensity of $CH_4$ (i.e., Tracer concentration of $CH_4$) at any given location once our final classifier model detects the presence of $CH_4$ at that location. As mentioned before, our regression models for $CH_4$ intensity prediction were trained on ten meteorological parameters, and all these features were standardized before fitting into multiple models for regression. For the performance evaluation of our regression models, we used the model with the best $R^2$ value on the validation set. Later we validated our results with $R^2$ score, Mean Square Error (MSE) and Root Mean Square Error (RMSE) score on the test data using the following equation,

$$R^2 = 1 - \frac{SSR}{SST}$$

$$\text{MSE} = \frac{1}{N}\sum_{i=1}^{N} SST$$

$$RMSE = \sqrt{MSE} = \sqrt{\frac{1}{N}\sum_{i=1}^{N} SST}$$

where, **SSR** refers to the sum of squared differences (residuals) between the actual values and the predicted values, **SST** refers to the sum of squared distances of the actual values from the mean of the actual values(totals), and **N** equals to the number of samples.

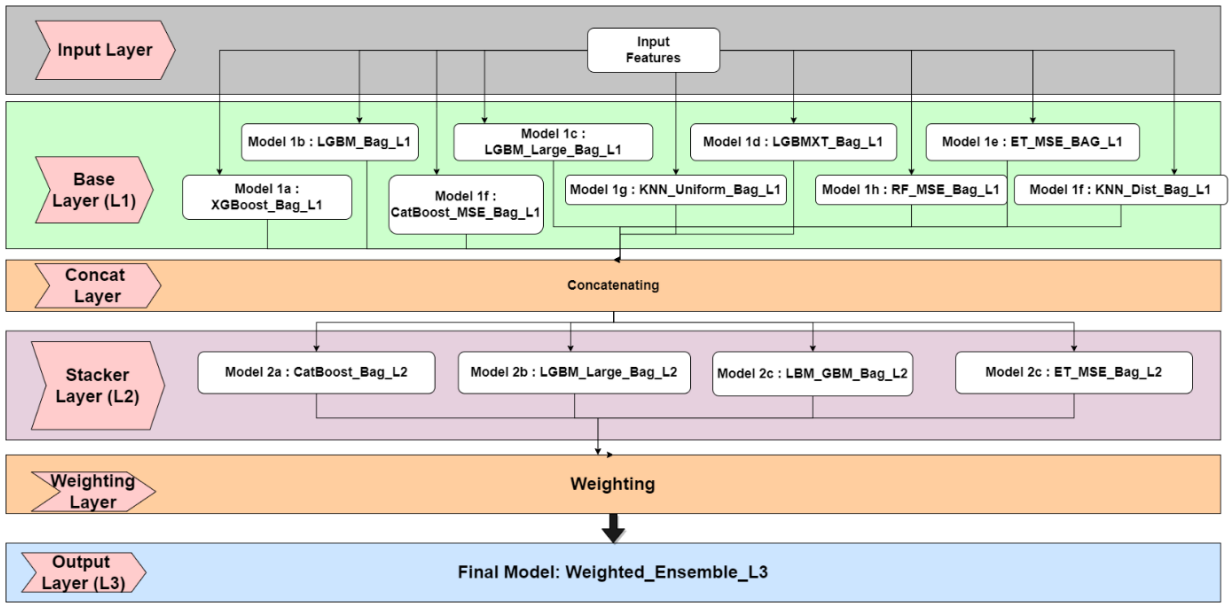

Fig. 4. Steps and weaker regression models involved in generating the best-performing regression models using ensemble learning

TABLE III. PERFORMANCE OF REGRESSION MODELS ON TEST SET

| Model | Performance Metric | | | |
| --- | --- | --- | --- | --- |
| | Root Mean Squared Error (RMSE) | Mean Squared Error (MSE) | Testing Set R2 | Validation Set R2 |
| WeightedEnsemble_L5 | 0.388 | 0.150 | 0.856 | 0.804 |
| CatBoost_BAG_L4 | 0.382 | 0.146 | 0.860 | 0.800 |
| XGBoost_BAG_L4 | 0.387 | 0.150 | 0.856 | 0.779 |
| LGBM_XT_BAG_L4 | 0.392 | 0.153 | 0.853 | 0.780 |
| ET_MSE_BAG_L4 | 0.394 | 0.156 | 0.851 | 0.797 |
| RF_MSE_BAG_L4 | 0.397 | 0.158 | 0.848 | 0.784 |
| LGBM_BAG_L4 | 0.398 | 0.158 | 0.848 | 0.791 |

| Model | Performance Metric | | | |
|---|---|---|---|---|
| | Root Mean Squared Error (RMSE) | Mean Squared Error (MSE) | Testing Set R2 | Validation Set R2 |
| LGBM_Large_BAG_L4 | 0.408 | 0.167 | 0.840 | 0.790 |
| WeightedEnsemble_L4 | 0.375 | 0.140 | 0.865 | 0.803 |
| XGBoost_BAG_L3 | 0.366 | 0.134 | 0.871 | 0.783 |
| RF_MSE_BAG_L3 | 0.369 | 0.136 | 0.869 | 0.791 |
| CatBoost_BAG_L3 | 0.377 | 0.142 | 0.864 | 0.801 |
| ET_MSE_BAG_L3 | 0.378 | 0.143 | 0.863 | 0.798 |
| LGBM_XT_BAG_L3 | 0.384 | 0.147 | 0.858 | 0.785 |
| LGBM_BAG_L3 | 0.385 | 0.148 | 0.858 | 0.781 |
| LGBM_Large_BAG_L3 | 0.397 | 0.158 | 0.848 | 0.770 |
| **WeightedEnsemble_L3** | **0.385** | **0.148** | **0.858** | **0.812** |
| XGBoost_BAG_L2 | 0.376 | 0.142 | 0.864 | 0.784 |
| RF_MSE_BAG_L2 | 0.381 | 0.145 | 0.860 | 0.801 |
| ET_MSE_BAG_L2 | 0.383 | 0.147 | 0.859 | 0.807 |
| LGBM_Large_BAG_L2 | 0.386 | 0.149 | 0.857 | 0.787 |
| CatBoost_BAG_L2 | 0.390 | 0.152 | 0.854 | 0.808 |
| LGBM_XT_BAG_L2 | 0.400 | 0.160 | 0.847 | 0.788 |
| LGBM_BAG_L2 | 0.405 | 0.164 | 0.843 | 0.799 |
| WeightedEnsemble_L2 | 0.437 | 0.191 | 0.816 | 0.792 |
| CatBoost_BAG_L1 | 0.417 | 0.174 | 0.833 | 0.770 |
| LGBM_Large_BAG_L1 | 0.447 | 0.200 | 0.808 | 0.774 |
| XGBoost_BAG_L1 | 0.452 | 0.204 | 0.804 | 0.762 |
| ET_MSE_BAG_L1 | 0.459 | 0.211 | 0.798 | 0.766 |
| LGBM_BAG_L1 | 0.482 | 0.232 | 0.777 | 0.739 |
| RF_MSE_BAG_L1 | 0.494 | 0.244 | 0.766 | 0.771 |
| LGBM_XT_BAG_L1 | 0.502 | 0.252 | 0.758 | 0.720 |
| KNN_Distance_BAG_L1 | 0.607 | 0.368 | 0.647 | 0.577 |
| KNN_Uniform_BAG_L1 | 0.610 | 0.372 | 0.643 | 0.571 |

TABLE IV. HYPERPARAMETER SETTINGS OF THE FINAL MODEL FOR REGRESSION IN THIS STUDY

| Layer | Model | Number of Features | Hyperparameter |
|---|---|---|---|
| Output Layer (L3) | WeightedEnsemble_L3 | 4 | Ensemble Size: 100, LGBM_BAG_L2, CatBoost_BAG_L2, ET_MSE_BAG_L2, LGBM_Large_BAG_L2 |
| Stacker Layer (L2) | LGBM_BAG_L2 | 19 | Learning Rate: 0.05, Number of Boost Rounds: 58 |
| | CatBoost_BAG_L2 | 19 | Iterations: 714, Learning Rate: 0.05, Evaluation Metric: R2 |
| | ET_MSE_BAG_L2 | 19 | Number of Estimators: 300, Maximum Leaf Nodes: 15000, Criterion: Squared Error |
| | LGBM_Large_BAG_L2 | 19 | Learning Rate: 0.03, Number of Leaves: 128, Feature Fraction: 0.9, Minimum data in Leaf: 5, Number of Boost Rounds: 172 |
| Base Layer (L1) | RF_MSE_BAG_L1 | 10 | Number of Estimators: 300, Maximum Leaf Nodes: 15000, Criterion: Squared Error |
| | LGBM_XT_BAG_L1 | 10 | Learning Rate: 0.05, Extra Trees: True, Number of Boost Rounds: 3668 |
| | ET_MSE_BAG_L1 | 10 | Number of Estimators: 300, Maximum Leaf Nodes: 15000, Criterion: Squared Error |
| | KNN_Uniform_BAG_L1 | 10 | Weights: Uniform |
| | CatBoost_BAG_L1 | 10 | Iterations: 9226, Learning Rate: 0.05, Evaluation metric: R2 |
| | LGBM_Large_BAG_L1 | 10 | Learning Rate: 0.03, Number of Leaves: 128, Feature Fraction: 0.9, Minimum data in Leaf: 5, Number of Boost Rounds: 489 |
| | KNN_Distance_BAG_L1 | 10 | Weights: Distance |
| | LGBM_BAG_L1 | 10 | Learning Rate: 0.05, Number of Boost Rounds: 602 |

Our regression-based ensemble learning models were generated in two stages. In the first stage, our base layer (L1) models were trained directly on the ten meteorological parameters to predict the tracer concentration. These models include various configurations of LGBM, XGBoost, RF, ET, and KNN. In the second stage, we have the stacker layer (L2) models that were developed considering the ten meteorological parameters as well as the regression outputs from base layer regression models as input features to better predict the intensity of $CH_4$. Finally, weighted averaging was used to determine the optimum combination of stacker layer models that results in the best-performing model in the output layer (L3). Table III presents the performance of all our regression models from the base layer and stacker layer selected based on the highest $R^2$ scores on the validation and test set. Our final best-performing regression model is a weighted averaged ensemble model of four stacker layer models 2a to 2d (as shown in Figure 4), and it achieved an $R^2$ score of 0.812 on the validation set. Using the same model for the test set we received an $R^2$ of 0.858 which is higher than other models. We also calculated mean squared error (MSE) of 0.148 and Root Mean Squared Error of 0.385 which is lower than all other models with similar $R^2$ values (as shown in Table 3). Table IV provides the hyperparameter settings for all the models that were involved in generating our final best-performing regression model (i.e., Weighted_Ensemble_L3).

Figure 5 shows actual and predicted tracer concentrations from the best ensemble model at different locations using latitude and longitude along the x and y axes. We have presented the results for three different time periods. Data from these three time periods were not used during the machine learning training.

**Actual Data from Simulation**                **Predicted Data using Developed Model**

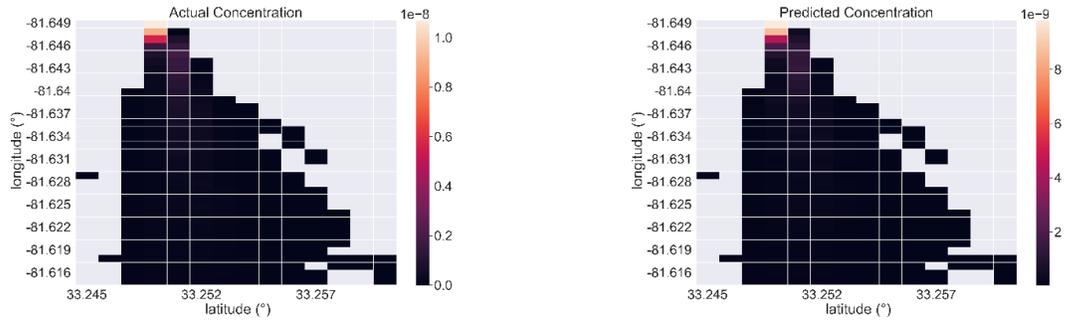

(a)

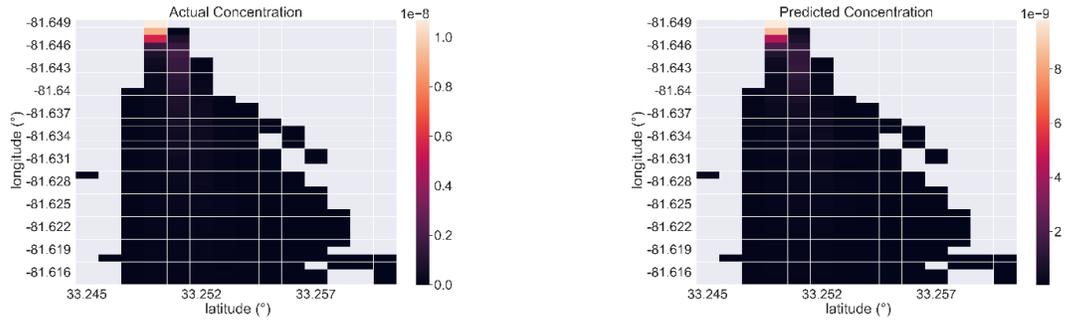

(b)

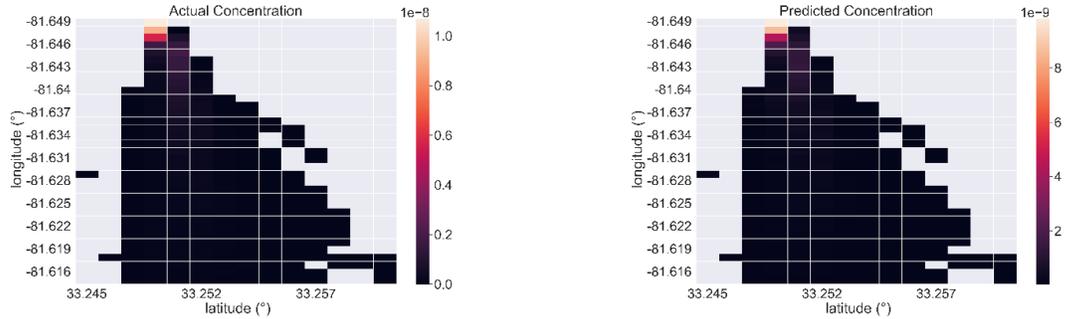

(c)

Fig. 5. Actual and predicted tracer concentration at (a) 5.00 PM, (b) 5.12 PM, and (c) 5.24 PM

### B. ML-Based Methane Detection in a Distributed System

The ML models used in this research to autonomously detect fugitive $CH_4$ emissions from environmental data and predict the corresponding intensity in the detected areas have been trained and validated using local computers in this study. However, such ML models can be extended into a distributed ML-based $CH_4$ environmental emission detection system by deploying multiple environmental sensing modalities, each possessing computational processing units (CPU).

Figure 6 displays the framework of a fugitive $CH_4$ environmental emission detection CPS where data storage, execution, optimization, and retraining of ML models will be accomplished by using cloud-based technology to ensure efficient real-time system operation. Additionally, real-time synchronous coordination of environmental data acquired by multiple sensor edges will achieve optimal system operation. The environmental emissions detection ML models will filter and calibrate meteorological information gathered from EPA ground stations as the UAVs monitor geographical areas of interest to keep false alarms minimal. As new environmental conditions are used to train the models, this distributed ML-based fugitive $CH_4$ environmental emissions detection CPS will achieve large-scale autonomous detection of $CH_4$.

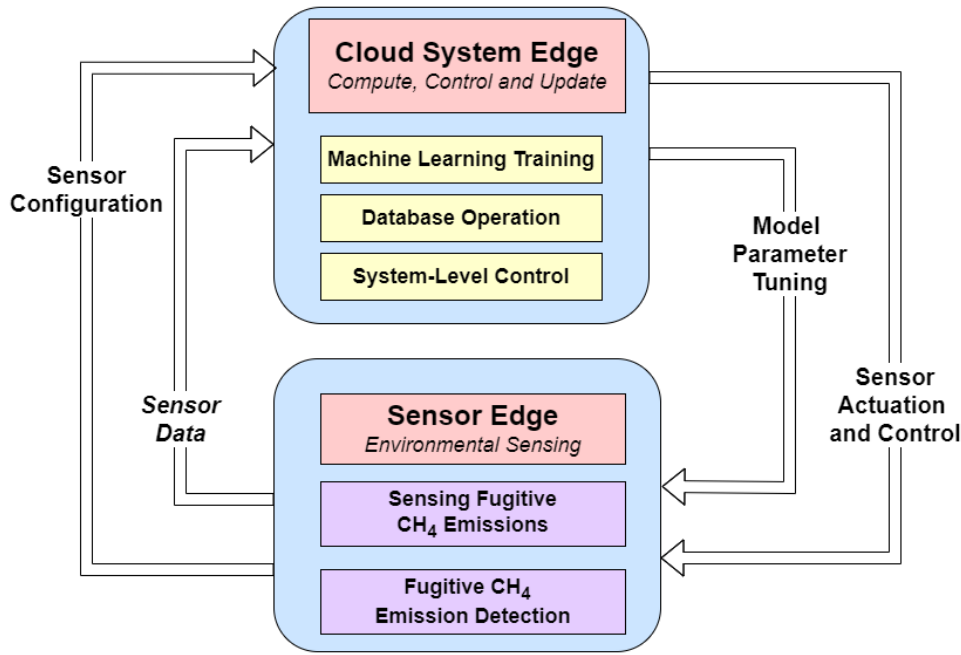

Fig. 6.  Fugitive $CH_4$ environmental sensing CPS

## V.  CONCLUSIONS

In this study, we developed an ensemble learning-based detection and intensity prediction system for environmental $CH_4$ emission. The dataset used in our analysis is a simulated dataset that was provided by the Savannah River National Laboratory. We formulated the $CH_4$ detection task as a classification problem and the $CH_4$ intensity prediction task as a regression problem. For both tasks, we used the ensemble learning method to generate ML models built upon several weaker ML models from base and stacker layers to yield better classification and regression performance. Our best-performing classifier model achieved an accuracy of 97.2% with an F1 score of 0.974, an MCC of 0.945, and an AUC ROC of 0.995 on the test dataset, whereas our best-performing regression model achieved an $R^2$ score of 0.858. One limitation of this study is that we utilized simulation data for developing our ensemble models. In the future, we will assess the performance of our model using real-world data to further validate the performance of our models in the detection and intensity prediction of environmental $CH_4$ emissions.

## VI.  FUTURE SCOPE

Gas dispersion models are computationally expensive and not suitable for real-time gas detection. This paper was our first step towards a scalable ML-based detection method using meteorological parameters and comparing it with dispersion models like HYSPLIT. In our next step, we will extend this research to gather meteorological data from real-world sensors on the edge to test and validate our ensemble learning models for detection.

One of the limitations of our model is that it is dependent on meteorological parameters only. In our next step, we plan to reduce the dependency on meteorological data by using and collecting $CH_4$ sensor data using a drone and generating sensor-specific ensemble learning models that can be deployed in the cloud. Multiple ML models for different data sources can work in parallel to better map a geographical area for $CH_4$ emissions.

## VII. Acknowledgements

This work was supported by the Department of Energy Minority Serving Institutions Partnership Program (MSIPP) managed by the Savannah River National Laboratory under BSRA contract TOA 0000525174 CN1. Any opinions, findings, conclusions, and recommendations expressed in this material are those of the author(s). They do not necessarily reflect the views of the Savannah River National Laboratory and the U.S. Government assumes no liability for the contents or use thereof.